 \documentclass[final,5p,times,twocolumn,authoryear]{elsarticle}

\usepackage{natbib}
\setcitestyle{numbers,square,comma}
\usepackage{lineno,hyperref}
\modulolinenumbers[5]

\usepackage{amssymb}
\usepackage{booktabs}
\usepackage{multirow} 
\usepackage{amsmath}  
\usepackage{subfigure}  
\usepackage{algorithm}
\usepackage{algorithmic}
\usepackage{color}
\usepackage{makecell}

\journal{Pattern Recognition}

\begin{document}

\begin{frontmatter}



\title{Conditional Pseudo-Supervised Contrast for Data-Free Knowledge Distillation}


\author[first]{Renrong Shao}
\author[first]{Wei Zhang\corref{mycorrespondingauthor}}
\ead{zhangwei.thu2011@gmail.com}
\fntext[]{This paper was published in Pattern Recognition 143 (2023) 109781. DOI:10.1016/j.patcog.2023.109781. ©2023 Elsevier Ltd. All rights reserved.}
\author[first]{Jun wang}
\affiliation[first]{
            organization={School of Computer Science and Technology, East China Normal Unversity},
            addressline={3663 North Zhongshan Road}, 
            city={Shanghai},
            postcode={200062}, 
            country={PR China.}
            }
\cortext[mycorrespondingauthor]{Corresponding authors.}
\begin{abstract}
Data-free knowledge distillation~(DFKD) is an effective manner to solve model compression and transmission restrictions while retaining privacy protection, which has attracted extensive attention in recent years.
Currently, the majority of existing methods utilize a generator to synthesize images to support the distillation. Although the current methods have achieved great success, there are still many issues to be explored. 
Firstly, the outstanding performance of supervised learning in deep learning drives us to explore a pseudo-supervised paradigm on DFKD.
Secondly, current synthesized methods cannot distinguish the distributions of different categories of samples, thus producing ambiguous samples that may lead to an incorrect evaluation by the teacher.
Besides, current methods cannot optimize the category-wise diversity samples, which will hinder the student model learning from diverse samples and further achieving better performance.
In this paper, to address the above limitations, we propose a novel learning paradigm, i.e., conditional pseudo-supervised contrast for data-free knowledge distillation~(CPSC-DFKD).
The primary innovations of CPSC-DFKD are: (1) introducing a conditional generative adversarial network to synthesize category-specific diverse images for pseudo-supervised learning, (2) improving the modules of the generator to distinguish the distributions of different categories, and (3) proposing pseudo-supervised contrastive learning based on teacher and student views to enhance diversity.
Comprehensive experiments on three commonly-used datasets validate the performance lift of both the student and generator brought by CPSC-DFKD.  The code is available at  https://github.com/RoryShao/CPSC-DFKD.git
\end{abstract}



\begin{keyword}
model compression\sep knowledge distillation\sep representation learning\sep contrastive learning\sep privacy protection



\end{keyword}

\end{frontmatter}



%

\section{Introduction}
\label{introduction}

With the development of artificial intelligence, the deep convolutional neural networks~(DCNNs) have been widely applied in various computer vision tasks and achieved remarkable success, such as image classification~\cite{krizhevsky2012imagenet}, object detection~\cite{ren2017faster}, and semantic segmentation~\cite{fu2019dual}.
Nevertheless, in practical applications, DCNNs suffer from some heavy issues.
Firstly, DCNNs always require heavy computation and storage. For example, only to handle one image, a VGG network commonly requires more than 500MB of memory, which makes them hard to be deployed on resource-constrained embedded or edge devices such as mobile phones and autonomous cars.
\begin{figure}[!ht]
    \vspace{-3mm}
    \centering
    \includegraphics[width=0.95\linewidth]{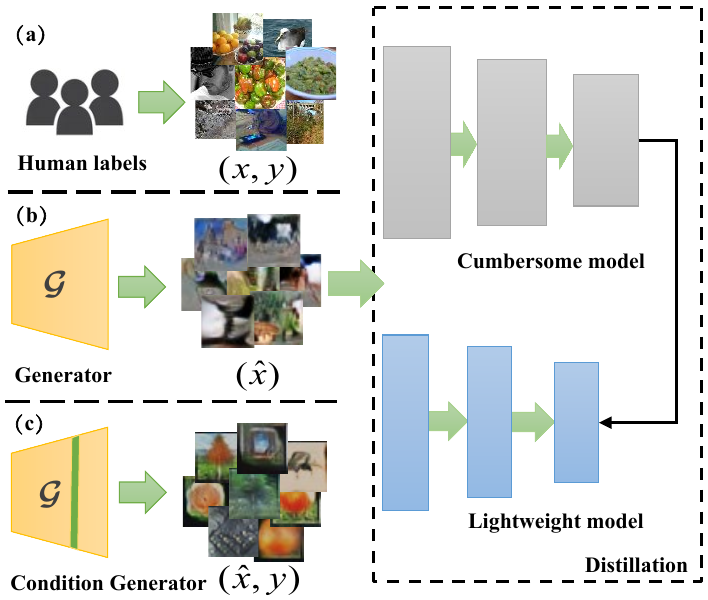}
    \vspace{-2mm}
    \caption{Conceptual diagram of different distillation approaches.~(a) Knowledge distillation by human labels.~(b) Previous data-free distillation approaches with the generator.~(c) Our proposed data-free approach by condition generator.}
    \label{fig:overview}
    \vspace{-4mm}
\end{figure}
Secondly, security and privacy concerns have aroused extensive attention. In some cases, training data cannot be publicly available, but pre-trained models are. For instance, the medical data or photos of users are usually sensitive and not publicly released. By contrast, the relevant pre-trained models do not involve user privacy and can be published.
Besides, the size and transmission are limited. The state-of-the-art models are usually trained on large datasets such as ImageNet~\cite{krizhevsky2012imagenet} with 14 million images, which is a heavy burden and a waste of resources to spread such large datasets.
To address the above issues, the data-free knowledge distillation~(DFKD) paradigm~\cite{chen2019data} is developed to compress the cumbersome deep models into more lightweight ones without access to or transferring the original training data.
This is opposed to the current knowledge distillation~(KD) paradigm shown in Fig.~\ref{fig:overview}~(a), which relies on source data and manual labeling.
It works by transferring the knowledge from a released pre-trained cumbersome teacher to a lightweight student without real data and aims to deploy the student instead of the teacher, as shown in Fig.~\ref{fig:overview}~(b).

Existing DFKD methods can be mainly divided into two paradigms.
The first paradigm is vanilla adversarial learning, in which the generator aims to synthesize discriminative samples according to the discrepancy between the teacher and student.
For example, Chen et al.~\cite{chen2019data} proposed the data-free learning method, which mainly exploits information entropy to optimize the generator and mimic the distribution of the teacher to optimize the student by cross entropy. Micaelli et al.~\cite{micaelli2019zero} mainly exploited adversarial distillation with Kullback-Leibler~(KL) divergence between the outputs of the teacher and student. This type of approach only utilizes the output layer of the teacher model for training, which is difficult to synthesize more realistic images and can easily lead to model collapse.
The second paradigm exploits the prior knowledge~(e.g., parameters of BatchNorm) in intermediate layers of the teacher to optimize the synthesized images, which is able to synthesize diverse images for training and mitigate model collapse.
For example, Yin et al.~\cite{yin2020dreaming} proposed to exploit Batch Normalization~(BN) to optimize the noise to generate images. Fang et al.~\cite{fang2021contrastive} proposed to employ BN to optimize the generator while improving the diversity of synthetic images by contrastive learning.
Although these methods improve the effect of synthetic images in some degree, there are still some limitations:
~(1)~In the data-free scenario, there is still a performance gap between teacher and student. This is in contrast to the data-driven scenario, where the performance of the student can even outperform the teacher.
(2) The current DFKD methods are all in an unsupervised manner due to the data-free scenario, while the most successful paradigm in deep learning is supervised learning which has surpassed human performance in some areas.
(3) Commonly, in existing methods, the generator widely exploits the vanilla BatchNormal layer to regularize the feature distribution of deconvolution, which is prone to make the learned distribution of categories focus on the dominant categories and ignore other categories, leading to ambiguous synthetic images.
(4) Although advanced methods~\cite{yin2020dreaming,fang2021contrastive} exploit the inversion to optimize diverse samples and aim to mitigate model collapse, they ignore to further distinguish the discrepancy under the category prior information, which may hinder the synthetic diversity and the effect of distillation.

To address the above limitations, this paper develops a novel learning paradigm, as depicted in Fig.~\ref{fig:overview}~(c), which is conditional pseudo-supervised contrast for data-free knowledge distillation, abbreviated as CPSC-DFKD.
The main innovations of CPSC-DFKD are three-fold:
(1) To utilize the idea of supervised learning, CPSC-DFKD introduces a conditional generator, which uses conditional category and random noise as the input of the generator to synthesize category-specific images. Consequently, the pseudo pairs of images with labels could be obtained, which enables student and generator learning in a pseudo-supervised manner.
(2) To distinguish and balance the distribution of different categories of samples, CPSC-DFKD improves the modules of the generator by categorical features embedding~(CFE) blocks, which connect the features and categories embedded in the intermediate layers.
Consequently, the generator can generate the different distributions of different categories by conditional information.
(3) To enhance the category diversity of synthesized images and the effect of distillation, CPSC-DFKD proposes to perform pseudo-supervised contrastive learning on the generator.
Different from the commonly-adopted data augmentation strategies, CPSC-DFKD exerts a unique teacher-student structure for image contrastive learning.
For a given image, the corresponding representations from the views of teacher and student form a positive pair in contrastive learning, while the representations of other images in the same batch provide negative signals.
Therefore, CPSC-DFKD is empowered by the representation uniformity of contrastive learning.

The main contributions of this paper are summarized as follows:
\begin{itemize}
    \item We propose a learning paradigm to improve DFKD with a conditional generative adversarial network, which is able to synthesize category-specific images and promote student learning.
    \item We introduce a categorical feature embedding block to effectively distinguish the distribution of different categories of samples, which connect the features and categories embedding in the middle layers.
    \item To our knowledge, we are the first to attempt to utilize the condition annotations to supervise the contrast of features representation of teacher and student in DFKD, which aims to optimize the diversity of synthesized images and improve the effect of distillation.
    \item Massive experiments are conducted on three mainstream benchmark datasets, {\it i.e.}, CIFAR-10, CIFAR-100, Tiny-imagenet.
    The results demonstrate the effectiveness of the proposed CPSC-DFKD in both improving the student and generator.
\end{itemize}

\section{Related Works}
\subsection{Generative Adversarial Networks}
GANs~\cite{goodfellow2014generative} establish a min-max game between a discriminator and a generator. The discriminator aims to distinguish generated data from real ones when the generator is dedicated to generating more realistic and indistinguishable samples to fool the discriminator. It has achieved great success in various image generation tasks, including image-to-image translation~\cite{chen2020distilling,YANG2022108208},
image super-resolution~\cite{QIAN2020107453,zhang2021data}, and image edit/inpainting~\cite{zhang2021pise, ZHANG2022108415}. 
Through adversarial training, GANs can synthesize fake images to support distillation between teacher and student~\cite{chen2019data,micaelli2019zero}.
However, vanilla GANs cannot synthesize images with specific categories and usually face some problems such as training instability and mode collapse. Therefore, there are many ways to improve the deficiencies of GANs, such as Conditional GANs~\cite{wang2018face}, Wasserstein GNAs~\cite{arjovsky2017wasserstein}, etc. Conditional GANs (CGANs)~\cite{wang2018face}, as another type of GANs, are proposed to utilize conditional information to generate images with labels for the discriminator, and they have been drawing attention as a promising tool for category-conditional image generation~\cite{isola2017image,li2020gan}.
In this work, we regard CGANs as one component of the proposed framework, which considers category-conditional information and synthesizes labeled images.

\subsection{Data-free Knowledge Distillation}
Knowledge distillation~(KD) is first proposed by~\cite{hinton2015distilling}, which aims to learn a compact and lightweight student model from the pre-trained powerful teacher model. To utilize more forms of knowledge to enhance the effect, many approaches have been proposed. For example, FitNet~\cite{romero2014fitnets} and AT~\cite{komodakis2017paying} transfer knowledge of hints and attention in the intermediate layers to the student, respectively. While RKD~\cite{park2019relational}, AMTML-KD~\cite{liu2020adaptive}, and RAD~\cite{liu2022coupleface} distill the relation-based knowledge of features, which can further improve the performance of the student. 
However, such approaches only focus on model compression.  
Currently, privacy preservation and data transmission restrictions
are gradually attracting attention, which leads to the emergence of research on DFKD.

DFKD methods incline to exploit a generator to synthesize massive samples to support the knowledge distillation learning between teacher and student~\cite{chen2019data,micaelli2019zero,fang2019data}.
Specifically, Chen et al.~\cite{chen2019data} proposed the DFAL framework which employs a generator to synthesize pseudo images and then makes student learn knowledge from teacher. Micaelli et al.~\cite{micaelli2019zero} exploited adversarial distillation to transfer the knowledge~(i.e., ZSKT) from teacher to student by  Kullback-Leibler~(KL) divergence and spatial attention.
And Fang et al.~\cite{fang2019data} proposed the DFAD framework which utilizes the MAE loss to fit the output distribution of the teacher.
However, this kind of method cannot synthesize realistic image and usually suffer from the risk of mode collapse~\cite{fang2021contrastive}.
Different from the above methods, Yin et al.~\cite{yin2020dreaming} utilized batch normalization statistics~(BNS) of teacher to normalize the noise, which can effectively synthesize more realistic samples for KD.
Similarly, such a trick is also introduced by DFQ~\cite{choi2020data} and CMI~\cite{fang2021contrastive} to optimize the images synthesized by the generator.
In DFQ, the generator is constrained to produce synthetic
images that match the original data distribution by referring to BNS from the batch normalization layers of teacher.
CMI also exploits BNS to incrementally synthesize some new samples by the contrastive diversity from a memory bank.
This category of methods can synthesize more realistic images to mitigate model collapse and achieve relatively better performance.

\subsection{Contrastive Learning}
Contrastive Learning has achieved a remarkable achievement in recent years~\cite{chen2020simple,he2020momentum,khosla2020supervised}.
The key to contrastive learning is to learn the effective embedding representations of data by deep models.
Normalized embeddings from the same class are pulled closer together, while embeddings from different classes are pushed away.
So far, there are some representative contrastive learning methods. They learn representations by maximizing the agreement between different representations of the same data example via a contrastive loss in hyperspace. For example, He et al.~\cite{he2020momentum} proposed MoCo to exploit contrastive learning as a dictionary look-up. Chen et al.~\cite{chen2020simple} proposed a simple framework~(i.e., SimCLR), which contrasts the representation without requiring specialized architectures or a memory bank. Different from the above unsupervised manners~\cite{he2020momentum,chen2020simple}, Khosla et al. proposed the SupCL~\cite{khosla2020supervised}, which introduces the labels for each representation and conducts the contrastive learning in a supervised manner.

Closely related to contrastive learning is the family of losses based on metric distance learning or triplets~\cite{WANG2022108589,qian2019softtriple}. 
These losses have been used to learn powerful representations, often in supervised settings, where labels are used to guide the choice of positive and negative pairs.
From a new aspect in this paper, we skillfully exploit the features of the samples extracted by teachers and students for comparison and utilize contrastive learning under supervised conditions to optimize the generator.
\begin{figure*}[!ht]
    \centering
    \vspace{-2mm}
    \includegraphics[width=\linewidth]{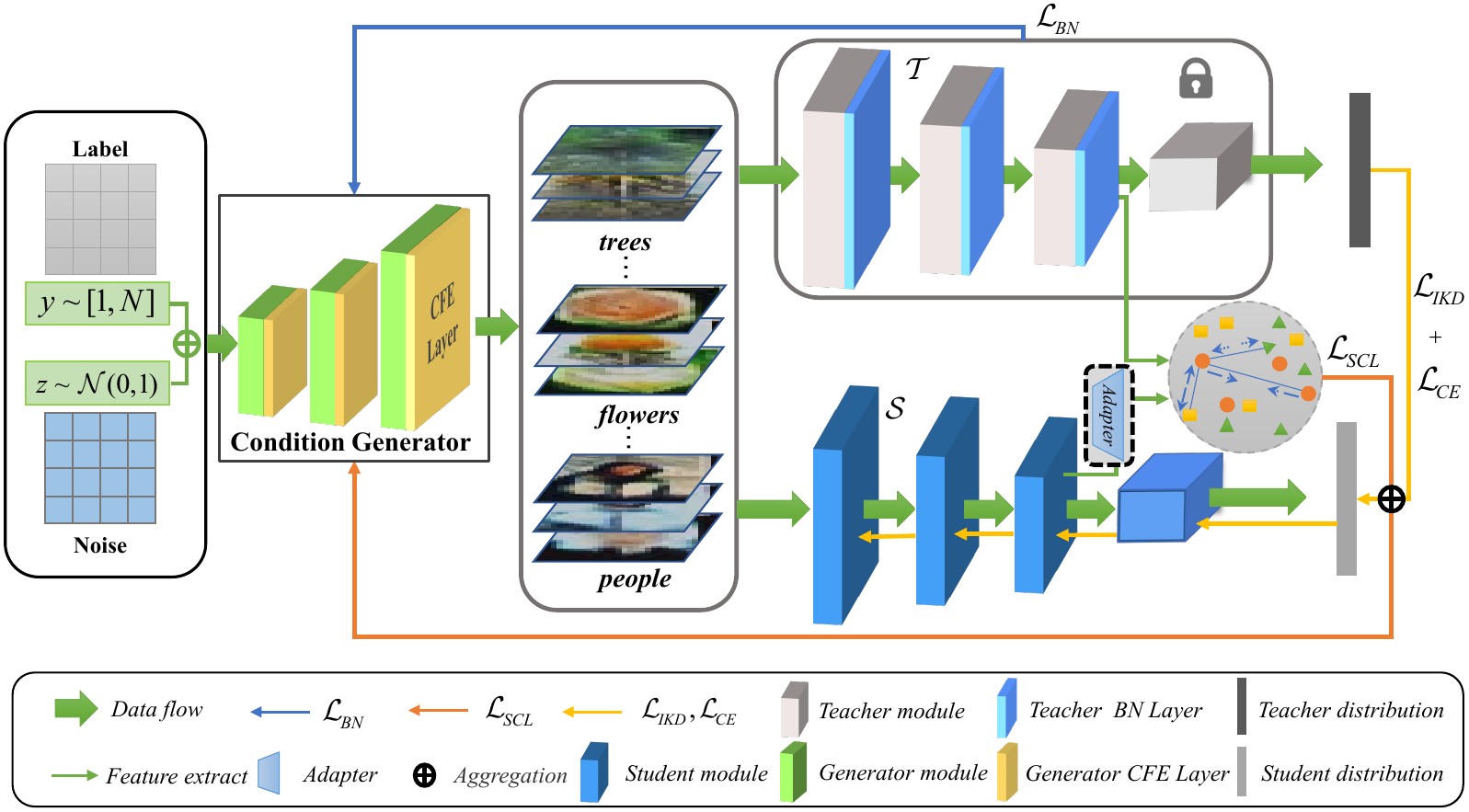}
    \vspace{-5mm}
    \caption{The overall workflow of CPSC-DFKD. A condition generator with CFE layers is exploited to synthesize images with labels for teacher and student by adversarial distillation. The CFE layer maps the category with features by rebuilding embedding layers based on BN layers.
    In the penultimate layer, we map the feature representation to a new space and compare the discrepancy under the supervision of category labels. Besides, distilling knowledge from teacher to student in the output layer.  }
    \label{fig:framework}
    \vspace{-5mm}
\end{figure*}

\section{Methods}
\subsection{Preliminaries}
DFKD is an effective paradigm to transfer knowledge from a cumbersome pre-trained teacher model $\mathcal{T}(x;\theta_{t}^{*})$  to a lightweight student model $\mathcal{S}(x;\theta_{s})$.
It works by minimizing the discrepancy $\mathcal{D}$ between teacher and student under the support of the synthetic image set ${\{\hat{x}_{i}|\hat{x}_{i}\in \mathcal{\hat{X}}\}}$, which expects to achieve a well-trained student model as follows:
\begin{small}
    \begin{equation}
        \theta _{s}^{*}= \mathop{\arg\min}\limits_{\theta_s}\mathbb{E}_{\hat{x}}[\mathcal{D}(\mathcal{T}(\hat{x};\theta_{t}^{*}), \mathcal{S}(\hat{x};\theta_{s}))] \,,
    \end{equation}
\end{small}
where $\mathcal{D}$ denotes the discrepancy metric,  e.g., KL divergence. 
$\mathcal{T}$ is pre-trained on a real dataset ${(x_{i},y_{i})\in \mathcal{X}}$ with the learned parameter $\theta_{t}^{*}$.
The synthetic dataset $\mathcal{\hat{X}}$ is used in the above equation since the real dataset is inaccessible due to privacy protection.
Existing studies commonly adopt the vanilla generator to synthesize alternative samples from noise $z\sim \mathcal{N}(0,1)$ for distillation~\cite{chen2019data,fang2019data}.
However, they neglect to distinguish the distribution of different categories of samples.
This issue is exacerbated when the amount of categories is large, for it is not conducive to generating specific images with category diversity.
To our investigation, there are very few works that employ a conditional generator for knowledge distillation, which motivates this paper to propose conditional adversarial distillation.

Besides, in the DFAD framework, the generator is more prone to model collapse due to inadequate optimization~\cite{chen2019data,micaelli2019zero,fang2019data}.
BNS is primarily introduced in~\cite{yin2020dreaming} and has also been demonstrated to be an effective manner to optimize the generator~\cite{fang2021contrastive, choi2020data}. It works by minimizing the distance between the current variance and mean of features and those of the pre-trained teacher in each batch normalization layer, just as follows:
\begin{small}
  \begin{align}
    \label{eq:BNS}
    \min_{\mathcal{G}} \left \{    \mathbb{E}_{\hat{x}} \left [  \sum_{l} \left \| \mu_{l} - \mu_{l}(\hat{x}) \right \|_{2} + \left \| \sigma_{l}^{2} - \sigma_{l}^{2}(\hat{x}) \right \|_{2} \right ]  \right \}  \,,
  \end{align}
\end{small}
where $\mu$ and $\sigma$ represent the mean and variance of BatchNormal, respectively. $l$ denotes the $l$-th BN layer of the teacher model.
With the BNS regularization, the synthetic images are more realistic, which is beneficial for distillation.

\subsection{Category-Conditional Generator}
The original DFKD only takes random noise as input and expects the generator to synthesize more realistic images to cover the original images.
In contrast, conditional adversarial distillation is devised to combine both the random noise ($z\sim \mathcal{N}(0, 1)$) and the conditional label set ($\{y_{i}| y_{i}\in \mathcal{Y}, i \in \mathrm{[1, N]}\}$, where $\mathrm{N}$ is the count of categories), as exhibited in Fig.\ref{fig:framework}.
The combined representations are taken as input to the generator $\mathcal{G}$.
Hence, we can control the synthetic images by the label information as conditions.

However, it is difficult for the generator to distinguish the distribution differences between different categories of samples during the min-max optimization.
To alleviate this issue, firstly, we exploit an improved conditional generator to synthesize category-wise samples to support distillation.
This is introduced in detail in the following.

Originally, in DCNNs, it is common to exploit a Batch Normalization~(BN) layer to normalize the distribution of features of convolution to reduce the internal covariate shift. Given a mini-batch $\mathcal{B}=\left \{ \mathrm{F}_{i,.,.,.} \right \}_{i=1}^{\mathrm{N}}$ of $\mathrm{N}$ samples, where $\mathrm{F}_{i} \in \mathbb{R}^{c \times h \times w}$. BN normalizes the feature maps at each layer as follows:
\begin{small}
\begin{align}
\label{eq:BNL}
\mathrm{BN}(\mathrm{F}_{i,c,h,w}|\gamma_{c}, \beta_{c} ) = \frac{ \mathrm{F}_{i,c,w,h}- \mathrm{E}_{\mathcal{B}}[\mathrm{F}.,c,.,.]}{\sqrt{\mathrm{Var}_{\mathcal{B}}[\mathrm{F}.,c,.,.]+\varepsilon }} \cdot \gamma_{c} + \beta_{c}  \,,
\end{align} 
\end{small}
where $\varepsilon $ denotes constant for numerical stability, $\mathrm{E}_{\mathcal{B}}$ and $\mathrm{Var}_{\mathcal{B}}$ are mean and variance of a batch, $\gamma_{c}$ and $\beta_{c}$ are trainable scalars corresponding to scale factor and bias, respectively.
However, the normalization of a batch of images has different categories. Theoretically, the distribution of different categories is dissimilar, so it is somehow not meticulous if BN is used directly.
It is not conducive for generative models to synthesize diverse samples of different styles.

Inspired by~\cite{miyato2018cgans,zhang2019self}, to ensure that the conditional generator can produce samples with diverse categories, we modify the generator model by introducing categorical features embeddings~(CFE) to modify the traditional BN layers.
Specifically, CFE replaces $\gamma_{c}$ and $\beta_{c}$ in Eq.~\ref{eq:BNL} by category-dependent feature maps $\mathbf{W}_{y} \in \mathbb{R}^{d \times m}$ and $\mathbf{b}_{y} \in \mathbb{R}^{d \times m} $.
$m$ is the dimension of a category and $d$ is the feature dimension corresponding to the category. The purpose of this improvement is to create a mapping between categories and features and learn the feature weights of different categories in the adversarial process.
The corresponding computational formula is defined as follows:
\begin{small}
\begin{align}
\label{eq:CFE}
\mathrm{CFE}(\mathrm{F}_{i,c,h,w}|\mathbf{W}_{y_{i}}, \mathbf{b}_{y_{i}} ) &=  \frac{ \mathrm{F}_{i,c,w,h}- \mathrm{E}_{\mathcal{B}}[\mathrm{F}.,c,.,.]}{\sqrt{\mathrm{Var}_{\mathcal{B}}[\mathrm{F}.,c,.,.]+\varepsilon }} \cdot \mathbf{W}_{y_{i}} + \mathbf{b}_{y_{i}}  \,,
\end{align}
\end{small}
where $y_{i} \in \mathcal{Y}$ indicates the category to which a sample belongs.
Based on this, the generator could make different image categories to learn more different latent distributions and synthesize category-wise diverse images.
They are beneficial for teacher and student to steadily support the distillation.

\subsection{Conditional Adversarial Distillation}
Adversarial distillation suggests training the model with a minimax game. During the stage of distillation training,~i.e., minimization stage, the student expects to mimic the teacher and minimize the discrepancy.
During the stage of adversarial training,~i.e., maximization stage, the generator anticipates generating as many real samples as possible while maximizing the possible discrepancy for a worst-case scenario.
The two training stages are alternated for convergence.
In order to adapt to the optimization of adversarial distillation, we introduce a conditional generator network $\mathcal{G}(z,y_{i}) \to (\hat{x}_{i},y_{i})$, where $\hat{x}_{i} \in \mathcal{X}$ and $y_{i} \in \mathcal{Y}$ to synthesize the alternative samples with conditional information to support the distillation, which is given by:
\begin{small}
\begin{align}
  \label{eq:gan}
    \min_{\mathcal{S}}\max_{\mathcal{G}} \left \{ \mathbb{E}_{(\hat{x},y)}\left [ \mathcal{D}(\mathcal{T}(\hat{x}|y),\mathcal{S}(\hat{x}|y) \right ]   \right \}  \,,
\end{align}
\end{small}
where $(\hat{x},y)$ is the synthetic images $\hat{x}$ accompanied by label $y$.
We omit the subscript $\hat{x}$ and $y$ for simplicity.
$\mathcal{D}$ measures the discrepancy between $\mathcal{T}$ and $\mathcal{S}$, which is commonly implemented by the KL  divergence~\cite{choi2020data,fang2021contrastive}.

To our investigation, when the number of categories is large, applying $l_{2}$-norm to further constrain the discrepancy between $\mathcal{T}$ and $\mathcal{S}$ can improve the effect of distillation  (as revealed in Tab.~\ref{tab:IKD_study}).
Because the KL loss focuses on the distribution of logits, when the sequence length of the categories is relatively large, the discrepancy of different categories is not always obvious.
Therefore, the whole discrepancy metric used in the approach is defined as follows:
\begin{small}
\begin{equation}
\label{eq:IKD}
 \mathcal{L}_{\mathrm{IKD}} = \min_{\mathcal{S}}\max_{\mathcal{G}} \left \{ \mathbb{E}_{(\hat{x},y)}  \ [\mathcal{D}_{\mathrm{KL}}(\mathcal{T}(\hat{x}|y), \mathcal{S}(\hat{x}|y)) + \alpha \mathcal{R}_{l_{2}}(\mathcal{T}(\hat{x}|y), \mathcal{S}(\hat{x}|y)) ] \right \} \,,
\end{equation}
\end{small}
where $\mathcal{R}_{l_{2}}$ denotes the logarithmic form of $l_{2}$-norm.
$\alpha$ is the hyperparameter to control the regularization loss. 

\subsection{Conditional Category and Distribution Alignment}
In the scenario of DFKD, conventional studies commonly assume that there is no exact label for supervision.
The prior study~\cite{chen2019data} makes an {\it one-hot} assumption about the network predictions on synthetic images. Differing from this study, we utilize the conditional category as the ground truth label for training both the generator and the student. For different goals, this approach can be divided into two stages:~(1)~\textbf{Maximization Stage.} Since the conditional pseudo labels are randomly initialized, there is some deviation between the conditional category and the true sample labels. For the teacher model, it already has the ability to discriminate the differences between different samples.
Therefore, we use the teacher model to discriminate samples with corresponding labels and minimize the difference between the output distribution and labels. This process forces the generator to generate samples that approximate the true samples which can be received by the teacher; 
~(2)~\textbf{Minimization Stage.} For the student, the conditional labels have been confirmed by the teacher model and thus can be used as real labels.
Therefore, cross-entropy losses are adopted to optimize the generator and student, which are given as follows:
\begin{small}
 \begin{equation}
 \label{eq:CE}
 \mathcal{L}_{\mathrm{CE}} =
 \left\{
 \begin{aligned}
   \mathbb{E}_{(\hat{x},y)} [ \min_{\mathcal{G}}  \frac{1}{n} \sum_{i=1}^{N}
   \mathcal{D}_{\mathrm{CE}}(\sigma(\mathcal{T}(\hat{x}|y)), y)  ]  ~~~   & \text{ if } stage~(1),
  \\
   \mathbb{E}_{(\hat{x},y)}  [ \min_{\mathcal{S}}  \frac{1}{n} \sum_{i=1}^{N}
   \mathcal{D}_{\mathrm{CE}}(\sigma(\mathcal{S}(\hat{x}|y)), y) ] ~~~   & \text{ if } stage~(2),
 \end{aligned}
  \right.
 \end{equation}
\end{small}
where $\sigma$ is the softmax function, $\mathcal{D}_{CE}(\cdot)$ is cross entropy,  $stage~(1)$ denotes the maximization stage, and $stage~(2)$ denotes the minimization stage.
 It is worth noting that the generator is optimized by the teacher but not by the student. This is intuitive since the teacher is well trained, being more stable and better to reflect the real data distributions than the student.

\subsection{Conditional Pseudo-Supervised Contrast}
\label{sec:supervised-contrast}

Contrastive learning~(CL) usually employs a model to extract feature representation and measure differences by pulling closer similar categories and pushing away dissimilar categories ~\cite{chen2020simple,khosla2020supervised}.
In DFKD, CL is first introduced in the study~\cite{fang2021contrastive}, which utilizes the teacher to extract local and global representations of synthetic images.
Different from the study~\cite{fang2021contrastive}, in our learning paradigm, both the teacher and the student are employed to extract different feature representations of the same synthetic image, respectively, then compare the representations under the supervision of a conditional category.
Moreover, CL is performed to benefit the training of the generator.
Through making the representations of different images to be uniformly distributed, it enhances the generator's capability of generating diverse images.

Specifically, for a batch of synthesized images $ \{(\hat{x}_{i}, y_{i})\}_{i=1}^N$, where $N$ is the batch size, the teacher and the student generate two representations for an image $\hat{x}_{i}$ by
$z_{s}^{i} = f_{s}(\hat{x}_{i}|y_{i})$ and $z_{t}^{i} = f_{t}(\hat{x}_{i}|y_{i})$.
$f_{t}(\cdot)$ is representation network of the teacher $\mathcal{T}$ without last layer. %
And it is analogous to $f_{s}(\cdot)$ of the student.
We assume that the image representation from the teacher as the anchor sample (e.g., $z_{t}^{i}$) and the corresponding image representation from the student (e.g., $z_{s}^{i}$) as a positive one.
The rest of the $\mathrm{N}-1$ image representations from the student in the same batch are taken as negative ones.
Consequently, we define the contrastive learning loss as follows:
\begin{small}
    \begin{equation}
      \label{eq:SCL}
      \mathcal{L}_{\mathrm{SCL}}=\mathbb{E}_{(\hat{x},y)} \left \{ \max_{\mathcal{G}} \left [ \sum_{i=1}^{N} \frac{-1}{\left |\mathcal{P}(i)\right | }  \sum_{y_{i} \in \mathcal{P}(i)} \log \frac{\exp (z_{t}^{i} \cdot z_{s}^{i} / \tau)}{\sum_{j, i \ne j}^{N} \exp ( z_{t}^{i} \cdot z_{s}^{j}/ \tau)} \right ]  \right \}  \,,
    \end{equation}
\end{small}
where the $\cdot $ sign denotes the inner (dot) product.
$i$ and $j$ are the indices of the samples in a batch.
$\tau$ denotes a temperature parameter, $\mathcal{P}(i) \equiv \left \{\forall y_{i} \in \mathcal{Y}: i \in \mathrm{N} \right \} $ is the set of conditional annotations corresponding to positive samples, which guides the sum of same categories in a batch. $|\mathcal{P}(i)|$ is the cardinality.
Since this loss could maximize the representations from different categories, it encourages the generator to closely aligned distribution to all entries from the same class.

\renewcommand\arraystretch{1} %
\newcolumntype{Y}{>{\centering\arraybackslash}X}

\begin{table*}[htbp]
  \centering
  \caption{Comparison of different data-free distillation approaches on CIFAR-10, CIFAR-100, and Tiny-ImageNet.}
  \vspace{2mm}
  \resizebox{0.95\linewidth}{!}{
    \begin{tabular}{c|cccc|cccc|cccc}
    \Xhline{1.1pt}
     Datasets & \multicolumn{4}{c|}{CIFAR-10} & \multicolumn{4}{c|}{CIFAR-100 } & \multicolumn{4}{c}{Tiny-ImageNet}  \\
    \hline
    Teacher & ResNet-34 & VGG-11 & WRN-40-2 & & ResNet-34 & VGG-11 &  WRN-40-2 & & ResNet-34 & VGG-11& WRN-40-2 & \\
   FLOPs &  74.9M &  272.5M &  329.0M & &  75.0M &  272.9M &  329.0M & &  300.0M &  732.0M &  1.3G & \\
    Params &  21.3M  &   128.8M  &  2.2M & &   21.3M  &   129.2M  &   2.3M &  &   21.4M  &   129.6M  &   2.3M & \\
    Acc. & 95.70  & 92.25 & 94.87 & & 78.05 & 71.32  & 75.83 & & 67.21 &  53.67  &   60.65 & \\
    \hline
    Student & ResNet-18 & ResNet-18 & WRN-16-2 & & ResNet-18 & ResNet-18 & WRN-16-2 & & ResNet-18 &  ResNet-18  &  WRN-16-2 & \\
     FLOPs  &   37.1M  &   37.1M  &   101.6M & &   37.2M  &   37.2M  &   101.6M  &  &   148.6M  &  148.6M  &   406.4M  & \\
     Params  &   11.2M  &   11.2M  &  0.7M &  &   11.2M  &   11.2M  &   0.7M &    &   11.3M  &   11.3M   &   0.7M &  \\
    Acc. & 95.35  & 95.35  & 93.95 &   & 77.10  & 77.10  & 69.87 &   & 64.53 &  64.53 &  56.44 &  \\
    \hline
    DFAL~\cite{chen2019data}  & 92.22 & 81.10 & 81.55 &  & 74.47 & 57.29 & 40.00 &  &  N/A  &  N/A  &   N/A &   \\
    ZSKT~\cite{micaelli2019zero}  & 93.32 & 89.46 & 89.66  & & 67.74 & 34.72 & 28.44 &  & N/A &   N/A  &  N/A &   \\
    DFAD~\cite{fang2019data}  & 93.30 & 90.90 & 91.42 &  & 67.73 & 55.93 & 35.01 &  & 57.34 & 38.30  &  33.07 & \\
    ADI~\cite{yin2020dreaming}   & 93.26 & 90.36 & 89.72 &  & 61.32 & 54.13 & 61.34 & & 60.21 &  46.67  & 45.22 & \\
    DFQ~\cite{choi2020data}   & 94.61 & 90.84 & 92.01 &  & 77.01 & 68.32 & 59.01 & & 62.44 & 48.33 & 47.63 &\\
    CMI~\cite{fang2021contrastive}   & 94.81 & 91.13 & 92.52 &  & 77.04 & 70.56 & 68.72 & & 63.91 & 51.41 & 51.64 & \\
    \hline
    Ours  & \textbf{95.33} & \textbf{92.16} & \textbf{93.35} &  &\textbf{77.39} & \textbf{71.56} & \textbf{69.03} &  & \textbf{64.57} & \textbf{53.25} & \textbf{53.72} & \\
    \Xhline{1.1pt}
    \end{tabular}%
    }
  \label{tab:main}%
\end{table*}%

\subsection{Objective Losses for CPSC-DFKD}
\label{sec:optimization}
In CPSC-DFKD, the major aim is to learn the student and the generator based on a well-trained teacher.
The objective loss for the student is given as follows:
\begin{small}
    \begin{equation}
    \label{eq:mins}
    \min_{\mathcal{S}} \mathcal{L}_{\mathrm{IKD}} + \eta \mathcal{L}_{\mathrm{CE}}\,.
    \end{equation}
\end{small}
Thus the student is trained to simultaneously mimic the output of the teacher and the conditional labels.

For the generator, we combine the $\mathcal{L}_{\mathrm{IKD}}$, $\mathcal{L}_{\mathrm{CE}}$, $\mathcal{L}_{\mathrm{SCL}}$, and $\mathcal{L}_{\mathrm{BN}}$ (similar as Eq.~\ref{eq:BNS}) losses.
Formally, the total objective loss for optimizing the generator is as follows:
\begin{small}
    \begin{equation}
    \label{eq:ming}
    \min_{\mathcal{G}}- \mathcal{L}_{\mathrm{IKD}}  + \beta   \mathcal{L}_{\mathrm{BN}} - \gamma  \mathcal{L}_{\mathrm{SCL}}  +  \eta \mathcal{L}_{\mathrm{CE}} \\,
    \end{equation}
\end{small}
where $\beta$, $\gamma$, $\eta$ are the hyperparameters to control the relative effect of each component.
Compared to the objective of the student, the generator's objective additionally contains the contrastive learning loss and the BNS regularization loss.
As aforementioned, the contrastive learning loss encourages the generator to synthesize diverse images.
And the intuition behind using the BNS loss is to make the synthetic image more realistic.
We detail the whole training pipeline in Algorithm~\ref{alg:1}, which mainly consists of the alternated student training stage and generator training stage.
\begin{algorithm}[!ht]
\caption{Training Algorithm for CPSC-DFKD.}
\label{alg:1}
\textbf{Input}: A pre-trianed teacher $\mathcal{T}$ on real data, generator $\mathcal{G}$ and student $\mathcal{S}$. \\
\textbf{Output}: A lightweight student $\mathcal{S}$, and a conditional generator $\mathcal{G}$. 

\begin{algorithmic}[1]
\FOR{\textit{number of iterations}}
\STATE \textbf{//  Minimization Stage}\\
\FOR{\textit{k steps iterations}}
\STATE Generate random noise $z \sim \mathcal{N}(0,1)$ and random categories condition $y \sim [0, \mathrm{N-1}]$; \\
\STATE Synthesize categorical image $\hat{x} = \mathcal{G}(z|y)$ ; \\
\STATE Calculate distribution discrepancy by $\mathcal{L}_{\mathrm{CE}}$ and $\mathcal{L}_{\mathrm{IKD}}$; \\
\STATE Froze $\mathcal{G}$ and $\mathcal{T}$, update $\mathcal{S}$ by Eq.~\ref{eq:mins} ; \\
\ENDFOR
\STATE \textbf{// Maximization Stage} \\
\STATE Generate random noise $z \sim \mathcal{N}(0,1)$ and random categories condition $y \sim [0, \mathrm{N-1}]$; \\
\STATE Synthesize categorical image $\hat{x} = \mathcal{G}(z|y)$ ; \\
\STATE Evaluate distribution discrepancy by $\mathcal{L}_{\mathrm{CE}}$ and  $\mathcal{L}_{\mathrm{IKD}}$; \\
\STATE Optimize diversity of the generator by   $\mathcal{L}_{\mathrm{BNS}}$ and  $\mathcal{L}_{\mathrm{SCL}}$;\\
\STATE Froze $\mathcal{S}$ and $\mathcal{T}$, update $\mathcal{G}$ by Eq.~\ref{eq:ming}. \\
\ENDFOR
\end{algorithmic}
\end{algorithm}

\begin{table*}[htbp]
  \centering
  \vspace{-2mm}
  \caption{Performance comparison with heterogeneous teacher and student models.}
  \vspace{2mm}
   \resizebox{0.8\linewidth}{!}{
    \begin{tabular}{cccccccccc}
    \toprule
    Datasets & \multicolumn{1}{c}{T.} & \multicolumn{1}{c}{S.} & DAFL~\cite{chen2019data}  & ZSKT~\cite{micaelli2019zero} & DFAD~\cite{fang2019data} & ADI~\cite{yin2020dreaming}   & DFQ~\cite{choi2020data}   & CMI~\cite{fang2021contrastive}   & Ours \\
    \midrule
    \multicolumn{10}{c}{WRN-40-2~(T.) and  WRN-40-1~(S.) } \\
    \midrule
    CIFAR-10 & 94.87 & 93.95 & 81.55 & 89.66 &  89.96 &  89.72 & 92.01 & 92.52 & \textbf{93.39} \\
    CIFAR-100 & 75.83 & 72.19 & 34.66 & 29.73 & 58.47 & 61.33 & 61.92 & 68.88 & \textbf{69.14 }\\
     Tiny-ImageNet  &  60.65   &  56.90  &  N/A  &  N/A &  34.03  &  45.90  & 49.37  & 51.16  &  \textbf{52.13}  \\
    \midrule
    \multicolumn{10}{c}{ ResNet-18~(T.) and  MobileNet-v2~(S.)} \\
    \midrule
    CIFAR-10 & 95.35 & 90.98 & 80.12 & 84.45 &  85.73 &  86.81 & 87.67 & 88.41  & \textbf{89.63} \\
    CIFAR-100 & 77.10 & 68.38 & 47.81 & 46.65 & 48.31 & 51.74 & 55.53 & 59.71 & \textbf{62.34}\\
     Tiny-ImageNet  &  64.53   & 55.06   &  N/A  &  N/A  &  35.61  &  41.83  & 44.28  & 47.85  &  \textbf{49.63 } \\
    \bottomrule
    \end{tabular}%
    \vspace{-4mm}
  \label{tab:diff_dimension}%
  }
  
\end{table*}%

\section{Experiments}
\subsection{Experimental Setup}
\textbf{Datasets.} We evaluate our algorithm on the image classification task on three widely used datasets, i.e.,  CIFAR-10, CIFAR-100, and Tiny-ImageNet.
The CIFAR-10 and CIFAR-100 datasets~\cite{alex2009cifar} are composed of 50,000 training and 10,000 testing images with 10 and 100 classes, respectively.
And the size of each image is 32$\times$32.
The TinyImageNet dataset~\cite{le2015tiny} contains 100,000 and 10,000 images from 200 object classes with their size 64$\times$64 for training and validation, respectively.
For both datasets, we perform the preprocessing of subtracting means and dividing by standard deviations in each RGB channel and employ the standard data augmentation techniques such as random cropping with zero padding and horizontal flipping.

\noindent\textbf{Models.} In our work, ResNet-34~\cite{he2016deep}, ResNet-18~\cite{he2016deep}, VGG-11~\cite{simonyan2014very}, and Wide ResNet-40-2~\cite{zagoruyko2016wide} are employed as the cumbersome teacher networks with pre-trained parameters for the three benchmark datasets, and other  five shallow networks, {\it i.e.}, ResNet-18~\cite{he2016deep}, MobileNet-v2~\cite{sandler2018mobilenetv2}, WRN-16-2~\cite{zagoruyko2016wide}, and WRN-40-1~\cite{zagoruyko2016wide} WRN-16-1~\cite{zagoruyko2016wide}, are utilized as the students.

\noindent\textbf{Generator architecture.} The generator in our approach is implemented by improved vanilla GANs, which synthesize images for CIFAR and Tiny-ImageNet, respectively.
The main architecture of our generators is shown in Tab.~\ref{tbl:gen-architecture}, where $\mathrm{Conv(3\times3)}$ denotes the $3\times3$ convolution kernels with stride $1\times1$ and the $\mathrm{CFE(Em_{w}, Em_{b})}$ layer denotes the categorical features embeddings layer with category-dependent feature maps.

\begin{table}[htbp]
  \centering
  \caption{Architecture of Generator.}\label{tbl:gen-architecture}
  \vspace{2mm}
  \resizebox{0.9\linewidth}{!}{
    \begin{tabular}{cc}
    \toprule

    conv\_blocks0 & FC, Reshape, $\mathrm{CFE(Em_{w},Em_{b})}$    \\

    conv\_blocks1 & $\mathrm{Conv({3\times3},1)}$, $\mathrm{CFE(Em_{w},Em_{b})}$, $\mathrm{LeakyReLU}$   \\

    conv\_blocks2 & $\mathrm{Conv}({3\times3},1)$, $\mathrm{CFE(Em_{w},Em_{b})}$, $\mathrm{LeakyReLU}$   \\

    conv\_blocks3 & $\mathrm{Conv({3\times3},1)}$, $\mathrm{Tanh}$, $\mathrm{CFE(Em_{w},Em_{b})}$   \\
    \bottomrule
    \end{tabular}%
    \vspace{-2mm}
  }
\end{table}%

\subsection{Implementation Details}
For fair comparisons with the other KD methods, we follow the CMI~\cite{fang2021contrastive} implementation environments using PyTorch. All the models are trained on NVIDIA 2080Ti GPUs with 12G memory.
We train the student with warm-up and optimize the parameters by the SGD optimizer with a momentum of 0.9, a batch size of 512 as default for CIFAR-10 and CIFAR-100, while the batch size of 256 and 64 for the Tiny-ImageNet and ImageNet, and the initial learning rate of student starts at 0.1 with a weight decay of 5 $\times 10^{-4}$.
We further use cosine annealing to update the learning rate within the total 600 epochs. 
However, if the batch size is too large, it will consume more memory, thus we utilize the mixed precision of float-16 to save memory while speeding up the training.
All the student models of our experiments are trained from scratch and evaluated by top-1 accuracy.
We exploit the full-convolutional generator following the architecture of GANs modified by replacing each BN layer with the categorical features embeddings layer.
Adam~\cite{kingma2014adam} is adopted to optimize the generator with an initial learning rate of $1 \times 10^{-3}$.
The learning rate of the generator is decayed by a factor of 10 at the 50-th and 150-th epochs.
To obtain the pre-trained teacher models, we train the teacher from scratch on each original dataset.
In order to facilitate comparison, we also utilize some pre-trained teacher models released by~\cite{fang2021contrastive}.
In the distillation stage, all the synthetic images have the same scale as the original input images in the pre-trained phase.

\begin{figure*}[!htbp]
  \centering
    \subfigure[Default other parameters as 1, adjust $\alpha$.]{
    \includegraphics[width=0.23\linewidth]{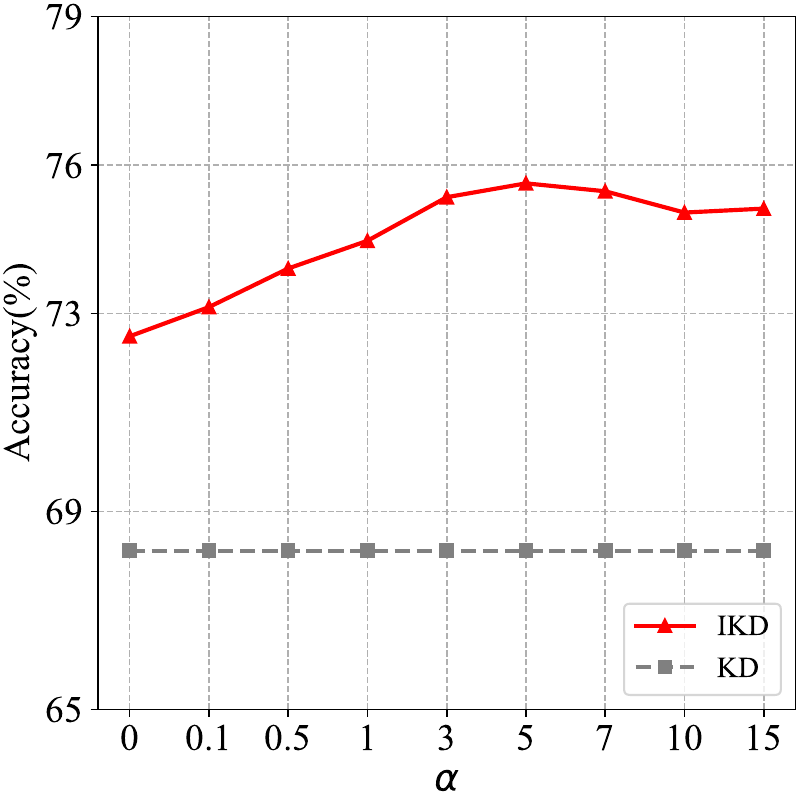}
    \label{fig:blocks_a}
     }
     \subfigure[Set $\alpha$ to 5, $\gamma$ and $\eta$ to 1, adjust $\beta$.]{
     \includegraphics[width=0.23\linewidth]{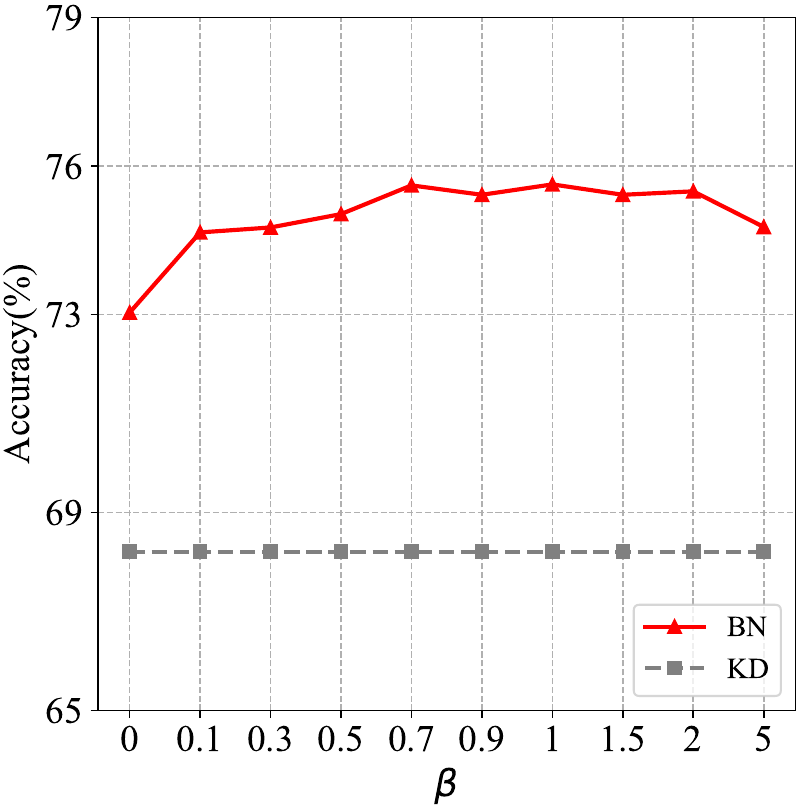}
     \label{fig:blocks_b}
     }
     \subfigure[Set $\alpha$ to 5, $\beta$ to 1 and $\eta$ to 1, adjust $\gamma$.]{
    \includegraphics[width=0.23\linewidth]{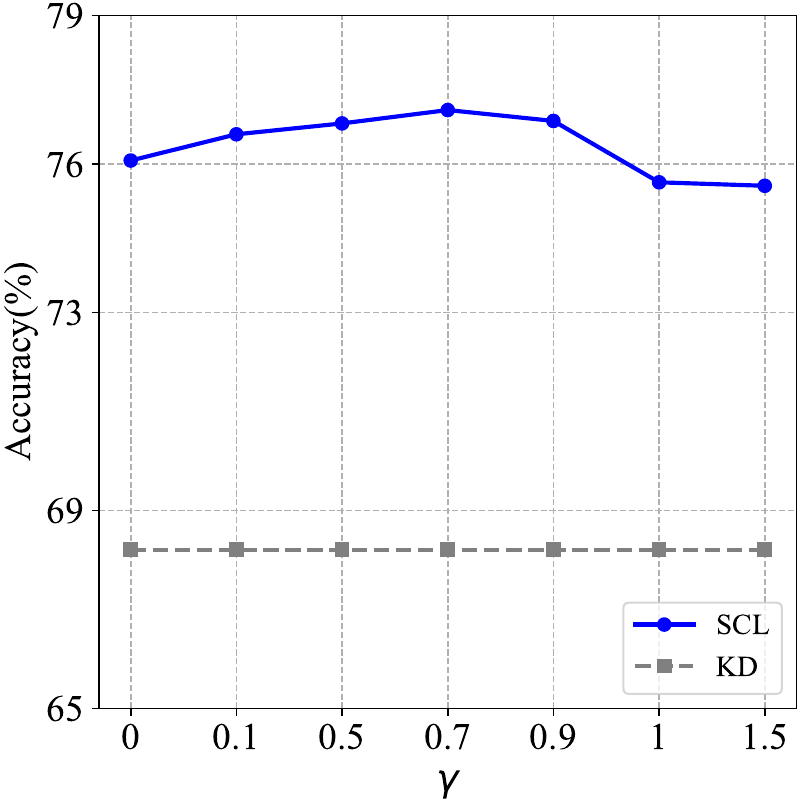}
    \label{fig:blocks_c}
     }
     \subfigure[Set $\alpha$ to 5, $\beta$ to 1 and $\gamma$ to 0.7, adjust $\eta$.]{
     \includegraphics[width=0.23\linewidth]{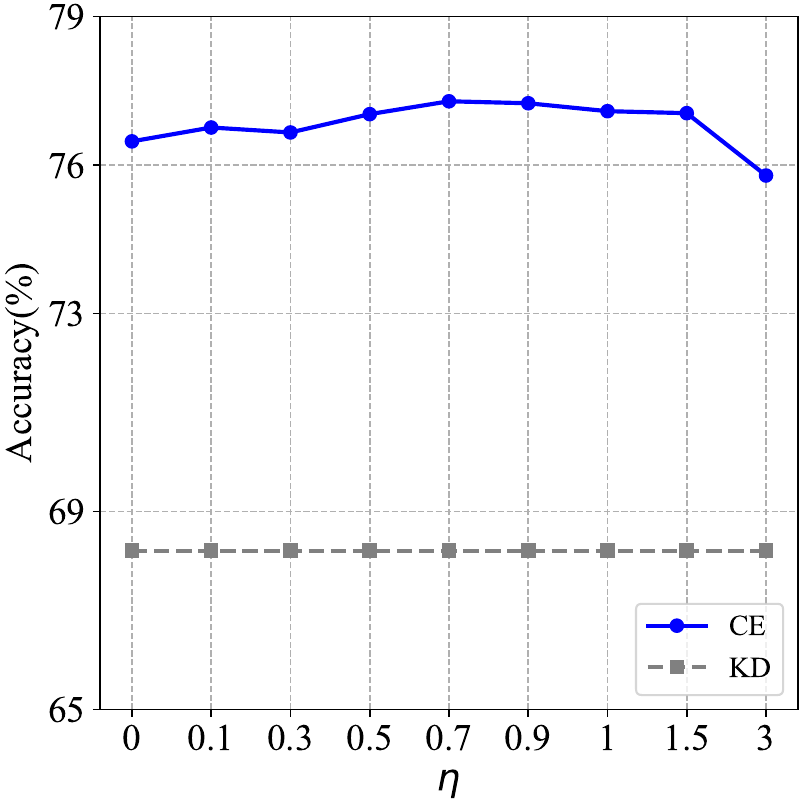}
     \label{fig:blocks_d}
     }
    \caption{Effect of $\alpha$, $\beta$, $\gamma$ and $\eta$ on the CIFAR-100 Dataset.}
  \vspace{-1mm}
  \label{fig:abcd}
\end{figure*}
\subsection{Comparison of DFKD Methods}

As exhibited in Tab.~\ref{tab:main}, we compare the proposed learning paradigm with the other representative approaches for data-free knowledge distillation, including DAFL~\cite{chen2019data}, ZSKT~\cite{micaelli2019zero}, DFQ~\cite{choi2020data}, DFAD~\cite{fang2019data},  ADI~\cite{yin2020dreaming}, and CMI~\cite{fang2021contrastive}.
The results are evaluated on the three widely used datasets with different scales, i.e., CIFAR-10, CIFAR-100, and Tiny-ImageNet.
The first two rows of the table depict the performance of the corresponding teachers and students trained on the   original real datasets from scratch, which are used as the reference.  It can be concluded from the table that DFKD can achieve model compression and computation acceleration without reducing model performance. For example, WRN-40-2 with WRN-16-2 can achieve more than 3$\times$ compression and acceleration on all datasets.

Basically speaking, the baselines could be divided into two categories.
(1) DAFL, ZSKT, and DFAD mainly imitate the teacher by reducing the discrepancy in the output layer.
Due to the limited diversity of synthetic images, it is apparent that they are inclined to suffer from model collapse, as shown in the first row of Fig.~\ref{fig:gen}.
(2) The other category of approaches, i.e., ADI, DFQ, and CMI, all utilize the pre-trained BNS to regularize the features of the teacher and student.
Therefore, this kind of approach is able to synthesize more realistic images with better diversity for distillation, as shown in the second row of Fig.~\ref{fig:gen}.
The proposed CPSC-DFKD employs the category-conditional generator to synthesize more diverse images, which performs better than the other baselines on the three datasets in most cases.
Besides, in the distillation stage, with the improved distillation and pseudo-supervision strategies, our approach can outperform the other baselines on the three datasets in most cases.
For example,  on CIFAR-10 and CIFAR-100 datasets, CPSC-DFKD improves the other baselines by 0.8 and 0.5 on average, respectively. 
On the Tiny-ImageNet dataset, CPSC-DFKD can outperform the other baselines by 1.5 on average. This proves that the conditional category method we introduced can successfully improve the effect of the model.

In the three group experiments, using VGG as the teacher model can significantly outperform the others.
The reason might be attributed to the fact that the original performance of student~(i.e., ResNet-18) is higher than teacher~(i.e., VGG-11), therefore, the student can
further improve, compared with the other two groups.
Besides, since we introduce the condition category as the pseudo supervision, the student model can be trained under the supervision of the pseudo labels. Therefore, compared to the other baselines,
ResNet-18~(VGG-11) can even outperform the performance of the teacher on the CIFAR-100 dataset.

Besides, to accommodate the contrastive learning of features in the penultimate layer between heterogeneous models, we have to address the dilemma of feature dimensional mismatch between the teacher and the student.
For example, the feature dimension extracted by WRN-40-2 is 128, while the dimension extracted by WRN-40-1 is 64.
As aforementioned, we introduce the $Adapter$ module which exploits the multi-layer perceptron with two hidden layers to map the feature dimension of the student to match the teacher.

As indicated in Tab.~\ref{tab:diff_dimension}, exploiting WRN-40-2 and WRN-40-1 on both CIFAR-10, CIFAR-100, and Tiny-ImageNet datasets, our proposed CPSC-DFKD approach can also outperform the advanced approach by 0.87\%, 0.26\%, 0.97\%, while the similar improvements are also performed by ResNet-18 and MobileNet-v2, proving that CPSC-DFKD is flexible to be adapted to different model architectures with the advantage of the $Adapter$ module.

\subsection{Ablation Study}
\subsubsection{Contribution of Loss}
\begin{table}[!htbp]
  \centering
  \vspace{-2mm}
  \caption{Effect of different components of CPSC-DFKD.}
  \vspace{2mm}
  \resizebox{0.9\linewidth}{!}{
    \begin{tabular}{ccc|ccc}
     \Xhline{1.1pt}
    $\mathcal{L}_{\mathrm{IKD}}$   & $\mathcal{L}_{\mathrm{SCL}}$  & $\mathcal{L}_{\mathrm{CE}}$  & CIFAR-10 & CIFAR-100 &  Tiny-ImageNet\\
    \hline
                      &                  &            & 92.44       & 68.09 & 49.63 \\
     $\checkmark$     &                  &                  & 94.03       & 69.94 & 51.04 \\
                      & $\checkmark$     &                  & 94.57       & 70.21 &  51.27 \\
                      &                   & $\checkmark$    & 93.77          & 68.53 & 50.82 \\
     $\checkmark$     & $\checkmark$     &                  & 95.16       & 70.40 & 52.33  \\
     $\checkmark$     &                  & $\checkmark$     & 95.06       & 70.07 & 53.06 \\
     $\checkmark$     & $\checkmark$     & $\checkmark$     & 95.33       & 71.56  & 53.72 \\
     \Xhline{1.1pt}
  \end{tabular}%
  \label{tab:ablation}%
  \vspace{-4mm}
  }
\end{table}%
The losses proposed in our approach are $\mathcal{L}_{\mathrm{SCL}}$, $\mathcal{L}_{\mathrm{CE}}$, and $\mathcal{L}_{\mathrm{IKD}}$, 
as revealed in Eq.~\ref{eq:mins} and \ref{eq:ming}.
To investigate the effect of each loss, we conduct an ablation study on CIFAR-10 (using ResNet-34 and ResNet-18), CIFAR-100 (using VGG-11 and ResNet-18), and Tiny-ImageNet~(using WRN-40-2 and WRN-16-2).
The detailed results are shown in Tab.~\ref{tab:ablation}, where we use $\checkmark$ to denote the corresponding loss used in training CPSC-DFKD.

As you can see, the first row shows our baseline accuracy on CIFAR-10, CIFAR-100, and Tiny-ImageNet of 92.44\%, 68.09\%, and 49.63\%, respectively, which exploits the  $\mathcal{D}_{\mathrm{KL}}$ to optimize the student while optimizing the generator by the $\mathcal{L}_{\mathrm{BN}} -\mathcal{D}_{\mathrm{KL}} $ loss.
In the second to fourth rows of the table are each of our methods combined with baseline loss separately, and it can be seen that our methods have different boosts on each of the two datasets. e.g., $\mathcal{L}_{\mathrm{SCL}}$ can boost the baseline to 94.57\%, 70.21\%, and 51.27\%, respectively, which demonstrates that an abundant and diverse sample is beneficial for enhancing the generalization of the model.
Furthermore, if we combine $\mathcal{L}_{\mathrm{IKD}}$ with any other two losses, the performance is largely improved.
This shows the necessity of fusing multiple losses.
We also compare their performance with the full approach and observe that all of the three variants are inferior to it, demonstrating that $\mathcal{L}_{\mathrm{SCL}}$, $\mathcal{L}_{\mathrm{CE}}$, and $\mathcal{L}_{\mathrm{IKD}}$ contribute positively to the final performance.

\subsubsection{Contribution of Components in Generator.}
Our approach leverages CFE to establish a mapping between the conditional category and the corresponding category distribution. Throughout the training process, the embedding layers of CFE acquire knowledge about the distribution information from the teacher, such as mean and variance, for each distinct category. To validate the efficacy of our approach, we conducted an additional study utilizing ResNet34 and ResNet-18 models.
The experimental results are shown in Tab.~\ref{tab:module_layer}.
As shown in the first row, when we adopt the standard BatchNormal layer instead of CFE to normalize the features for synthesizing images, the classification accuracy is 94.79\%, 76.32\%, and 52.33\% on CIFAR-10, CIFAR-100, and Tiny-ImageNet, respectively.
Compared with the final version 95.33\%, 77.39\%,  and 53.72\% of CPSC-DFKD, its performance is noticeably worse.
Moreover, we employ the generator with 3-layer CFE to train CPSC-DFKD.
The results show the performance is improved as compared to the former method, but still underperforms the final version of CPSC-DFKD.
This suggests that using full-layer CFE to generate different distributions of category-wise features is beneficial for distillation by the teacher.
\begin{table}[!htbp]
  \centering
  \caption{Comparison of different CFE layers.}
   \vspace{2mm}
   \resizebox{0.9\linewidth}{!}{
    \begin{tabular}{c|ccc}
    \Xhline{1.1pt}
    Generator Layer &    CIFAR-10 &    CIFAR-100 &      Tiny-ImageNet \\
    \hline
    w/o CFE & 94.79 & 76.32 &  52.33 \\
    \hline
    w/ 3-layer CFE & 95.01 & 76.74 & 53.04  \\
    \hline
    w/ full-layer CFE & 95.33 & 77.39 &  53.72 \\
    \Xhline{1.1pt}
    \end{tabular}%
    }
    \vspace{-4mm}
  \label{tab:module_layer}%
\end{table}%

\begin{figure*}[!ht]
  \centering
  \includegraphics[width=0.8\linewidth]{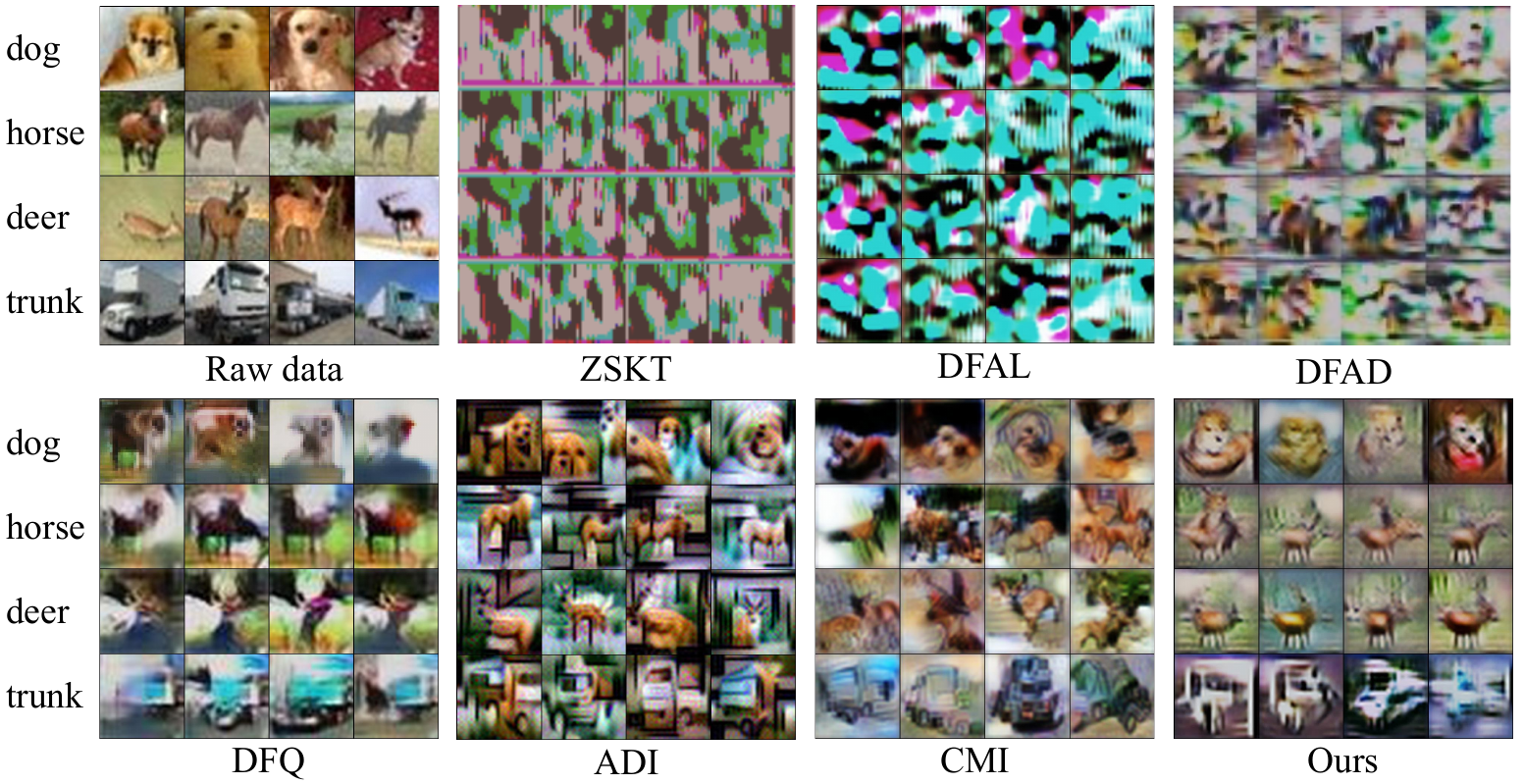}
  \caption{Visualization of synthetic category-specific diverse images from WRN40-2 to WRN16-1 by different approaches on CIFAR-10.}
  \vspace{-3mm}
  \label{fig:gen}
\end{figure*}
\subsubsection{Analysis of Improved Distillation.}
In previous distillation\cite{choi2020data,fang2021contrastive}, the teacher and the student mainly rely on KL divergence. According to our study, combining KL with L2 regularization can effectively improve the distillation effect. To verify the effectiveness of knowledge transfer, we conduct an ablation study~(with WRN-40-2 and WRN-16-2) on three datasets.
First, we only utilize the  $\mathcal{D}_{\mathrm{KL}}$ loss to transfer the knowledge and achieve the basic accuracy of 90.04\%, 66.34\%, and 45.08\% on three datasets, respectively. Then, we exploit the $\mathcal{R}_{l_{2}}$ loss without other strategies to promote the student mimic the distribution of teacher. As shown in Table~\ref{tab:IKD_study}, using $\mathcal{R}_{l_{2}}$ can outperform using $\mathcal{D}_{\mathrm{KL}}$ notably.
Finally, we combine the two strategies for training and find that the average experimental performance can be improved by about 2.5\%.
We argue that the KL loss focuses on the logit matching.
Nevertheless, the distribution between different categories is not always significant. Therefore, some approaches may introduce the temperature to soften the logits to mine the dark knowledge~(i.e., vanilla KD). While L2 loss can effectively scale and shrink the category differences, it is more conducive to distinguishing the differences between different categories and optimizing the model.

\begin{table}[!htbp]
  \centering
  \vspace{-2mm}
  \caption{ Performance of each component in improved distillation.} 
  \vspace{2mm}
  \resizebox{0.9\linewidth}{!}{
    \begin{tabular}{ccccc}
    \Xhline{1.1pt}
     Method  &   CIFAR-10  & CIFAR-100 & Tiny-ImageNet & Average \\
    \hline
       $\mathcal{D}_{\mathrm{KL}}$    &  90.04 &    66.34   &   45.08  &  67.15  \\
      $\mathcal{R}_{\mathrm{l}_{2}}$  &   91.41  &   67.20   &    47.35   &  68.65  \\
      $\mathcal{L}_{\mathrm{IKD}}$   &   92.29   &   67.73   &   48.83  &  69.62  \\
    \Xhline{1.1pt}
    \end{tabular}
  \label{tab:IKD_study}
  }
  \vspace{-5mm}
\end{table}

\subsubsection{Analysis of Hyperparameters.}

In the proposed method, there are four hyperparameters to control the whole optimization pipeline.
In this part, we utilize ResNet-34 and ResNet-18 on the middle-size dataset CIFAR-100 to analyze these hyperparameters.
We vary one hyperparameter and fix the other three hyperparameters We first change the hyperparameter $\alpha$ according to \{0, 0.1, 0.5, 1, 3, 5, 7, 10, 15\}, while the other three hyperparameters are set to 1.
It can be found when $\alpha$ takes the range of $\{3,7\}$, the performance is better, which is higher than the vanilla KD approach.
Then we analyze $\beta$ by tuning it according to \{0, 0.1, 0.3, 0.5, 0.7, 0.9, 1, 1.5, 2, 5\}.
As can be seen, when $\beta$ is in the range of $\{0.7,2\}$, better results could be achieved.
Then we analyze the other two hyperparameter losses which are supervised by pseudo labels. Similarly, we can observe that when they take suitable value ranges, better performance is achieved.

\subsection{Visualization Analysis and Evaluation }  %
\subsubsection{Analysis of Generator.}
\begin{figure*}[!ht]
  \centering
  \includegraphics[width=0.9\linewidth]{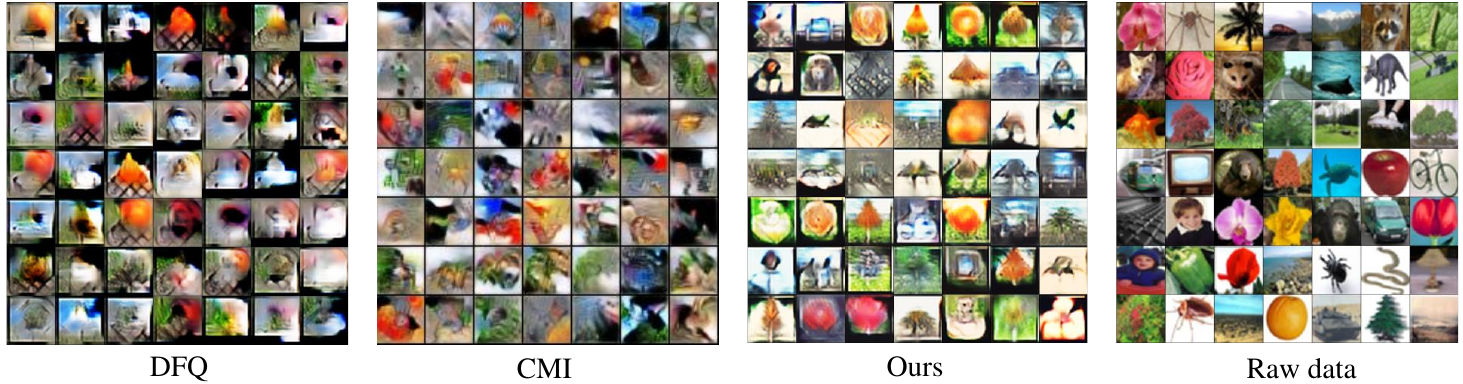}
  \caption{Visualization for synthetic samples from WRN-40-2 to WRN-16-1 by different approaches on CIFAR-100.}
  \label{fig:gen_cifar100}
\end{figure*}

In DFKD task, the generator is commonly exploited to support the distillation framework, which can synthesize the alternative samples. However, the generator is prone to model collapse in the training process,  making the model unable to generate diverse samples. This will affect the effect of the distillation~\cite{fang2021contrastive}. Therefore, in this part, we evaluate the quality of synthetic samples from two dimensions: visualization effect and generated sample score.
To analyze the effect of our generator, we visualize the synthetic images and compare them with other methods on the CIFAR-10 and CIFAR-100 datasets, respectively.
In Fig.~\ref{fig:gen}, we choose four categories~($dog$, $horse$, $deer$, $trunk$) of images synthesized by different approaches on the CIFAR-10 dataset.
The first column mainly shows the images synthesized by the first category of the DFKD methods ({\it{i.e.}},  DAFL, ZSKT, and DFAD).
Although the classification performance of these methods is not so bad, it is obvious that the synthetic images are so blurry that one can hardly distinguish the outline of objects.
The generators of these approaches incur model collapse due to inadequate optimization.
By contrast, the other category of approaches ({\it{i.e.}}, ADI, DFQ, and CMI) benefit from BNS and respective optimization, and thus they could synthesize more realistic and diverse images.
Likewise, with a conditional generator and corresponding optimization, our approach can synthesize more realistic images as well.
This is revealed by Fig.~\ref{fig:gen} where we can recognize different objects with various shapes.
Besides, we visualize the synthetic images of CIFAR-100 in Fig.~\ref{fig:gen_cifar100}. We can find the synthesis effect of DFQ and CMI is relatively obscure.  This might be because BN tends to learn some dominant categories without guiding them by category priors, which may result in ambiguous synthetic images. For example, the abundant synthetic images by DFQ take the keyboard incorrectly as the background.
We randomly select the generated images by the second category of the approaches.
For the synthetic images by CPSC-DFKD, it is clear to recognize different categories (i.e., $trees$, $flowers$, $people$) with different shapes, which illustrates the ability of our method to distinguish the difference in distribution between different categories.

For synthesized images, the Inception Score~(IS)~\cite{barratt2018note} and Frechet Inception Distance~(FID) score~\cite{heusel2017gans} are commonly used to evaluate the quality and diversity. However, the IS compares the synthesized images with the distribution of the Inception pre-trained model on ImageNet~\cite{deng2009imagenet}, which may lead to incorrect evaluation. While the FID compares the synthesized images and the original images by Inception, which is more suitable for our experiments, the formula is as follows:
\begin{small}
    \begin{equation}
      \label{eq:fid}
    \mathrm{FID}(x, \hat x) = \parallel \mu_{x} -\mu_{\hat x} \parallel+Tr(\sum_{x} + \sum_{\hat x}- 2(\sum_{x}\sum_{\hat x})^{\frac{1}{2} })
    \end{equation}
\end{small}
where $(x, \hat x)$ donates the original images and the images synthesized by GANs, $(\mu_{x}, \sum_{x})$ and  $(\mu_{\hat x}, \sum_{\hat x})$ are the mean and covariance statistics of original images and synthesize images. It is worth mentioning that a lower FID means that the generated distribution is closer to the real image distribution.
To this end, we utilize the FID score as the evaluation metric to compare our approach with CMI, DFQ, and ADI.
\begin{table}[htbp]
  \centering
  \vspace{-2mm}
  \caption{Evaluation of different approaches w.r.t. FID. }
  \vspace{2mm}
  \resizebox{0.6\linewidth}{!}{
    \begin{tabular}{ccc}
    \hline
    Methods  & CIFAR-10 & CIFAR-100  \\
    \hline
    ADI~\cite{yin2020dreaming}   & 93.42 & 98.54 \\
    DFQ~\cite{choi2020data}   & 82.36 & 89.72 \\
    CMI~\cite{fang2021contrastive}   & \textbf{77.93} &  84.34 \\
    Our   & 80.97 &  \textbf{79.99} \\
    \hline
    \end{tabular}%
    \vspace{-2mm}
  \label{tab:FID}%
  }
\end{table}%

Specifically, the 2048-dimensional features of the last pool layer of the inception net are extracted, which contain the semantic information of the synthetic images to a certain extent.
Consequently, we calculate the distance to estimate the quality image.
As can be seen from Tab.~\ref{tab:FID}, our performance is a bit lower than CMI on CIFAR-10 but outperforms ADI and DFQ. Algouth, the performance of synthetic images on CIFAR-10 is not very satisfactory, thanks to the improved knowledge distillation loss and pseudo supervision learning, our effect of distillation still outperforms the other methods.
On the CIFAR-100 dataset, our approach achieves state-of-the-art performance, which means that the images synthesized by our method are more diverse than others. Therefore, the distillation effect on the CIFAR-100 dataset is quite remarkable. We can find that when the category is large, the synthetic images are more effective, which demonstrates that the improved CFE layer in the generator can effectively learn different kinds of distribution information.

\subsection{Discussion About Other Compression Methods}
Essentially, our approach belongs to DFKD which is a special scenario in model compression. It can be used to address privacy protection and transmission limitations. 
Currently, there are some other mainstream methods of model compression, such as quantization~\cite{banner2018scalable, qin2020forward, qin2023distribution}, pruning~\cite{liurethinking,YAO2021108056} and so on. 
However, not every compression method works well in data-free scenarios. For this reason, we make a brief discussion as follows:
    \textbf{Pruning v.s. Distillation.} 
    Pruning~\cite{YAO2021108056, liurethinking} is a technique aimed at reducing the size and complexity of neural networks by removing network components that contribute little to the overall performance. This can be achieved through non-structure pruning such as weight and neuron pruning, or structure pruning including filter-wise~\cite{ YAO2021108056}, channel-wise~\cite{he2017channel}, stripe-wise pruning~\cite{liu2022soks}. However, pruning can potentially destroy the original network structure and hence requires iterative fine-tuning to maintain network performance. Additionally, the sparse rate, or pruning rate, needs to be considered in pruning.

    In comparison, distillation is a method that focuses on transferring knowledge from a larger and more complex model (teacher) to a smaller and simpler one (student), without breaking the network structure or requiring extensive training searches.
    As such, distillation is generally considered more convenient than pruning, which often relies on access to data for fine-tuning purposes and may not be applicable in data-free scenarios.

    \textbf{Quantization v.s. Distillation.}
    Model quantization~\cite{banner2018scalable} is essentially a technique for function mapping. It can be categorized into linear quantization and nonlinear quantization, depending on whether the mapping function is linear or not. Linear quantization, for instance, uses low-bit precision (e.g., 8-bit)~\cite{banner2018scalable} instead of high-bit precision (e.g., 32-bit). 
    One of the most effective quantization methods for achieving high compression rates is binary quantization~\cite{qin2020forward,qin2023distribution}, also known as 1-bit quantization. This unique method involves replacing the original float-32 values of weights or activation functions in the neural network with 1-bit values of either 0~(-1) or +1. Binary quantization significantly reduces model size and speeds up computation.
    However, the downside of this approach is that it often leads to a decrease in network performance, particularly on large datasets. To address this challenge, distillation can be employed~\cite{boo2021stochastic}, which not only compresses the smaller model but also improves the training effectiveness of the overall model. Besides, quantization can also be combined with data-free distillation~\cite{liu2021zero,choi2020data,xu2020generative}.

    \textbf{Other approaches.} In addition, low-rank approximation~\cite{lee2013local,lebedev2015speeding, kim2015compression} is a prevalent technique for achieving compression by sparsifying the convolutional kernel matrix by merging dimensions and imposing low-rank constraints. For example, singular value decomposition~(SVD)~\cite{lee2013local} commonly exploits two-dimensional matrix operations, while high-dimensional matrix operations often involve CP decomposition~\cite{lebedev2015speeding}, Tucker decomposition~\cite{kim2015compression}, etc. However, the decline in popularity of low-rank approximation can be attributed in part to the growing use of $1\times1$ convolutions in neural networks. The small size of these convolutions makes them difficult to accelerate and compress using matrix decomposition methods, which undermines the effectiveness of low-rank approximation. Furthermore, the performance of low-rank approximation tends to deteriorate when applied to larger networks, further limiting its appeal. Distillation, by contrast, is not limited to the structure of the model and is gaining popularity because of its flexibility and efficiency. 
    To the best of our knowledge, there has been a lack of research on the integration of low-rank approximation and data-free distillation techniques. This area presents promising opportunities for further exploration and investigation.

\section{Conclusion}
In this work, we have proposed Conditional Pseudo-Supervised Contrast for data-free knowledge distillation (CPSC-DFKD).
Thanks to three features, CPSC-DFKD enables better learning for both the student and the generator: (1) exploiting an improved conditional generator to synthesize the category-specific images for pseudo-supervised learning; (2) introducing an improved categorical feature embedding blocks to distinguish the different categorical distribution; (3) proposing contrastive learning to achieve diversity of synthesized images.
Extensive experiments have been conducted on three datasets and demonstrated that CPSC-DFKD can not only synthesize more categorical diverse images for the generator but also improve the performance of data-free knowledge distillation for the student.
Actually, our method can synthesize images with arbitrary data size by category.
Therefore, it can be further applied to the scenarios with category imbalance, such as long-tail learning. In addition, some other tasks such as semantic segmentation are also worthy of further exploration.
\subsection{Limitations}
However, in practice, the proposed approach might have the following limitations:
  
  \textbf{Heavy computation cost}. Although our method and existing methods have solved simple classification tasks, the training models from scratch might become difficult and costly when encountering complex tasks~(e.g., fine-grained vision classification, semantic segmentation, etc.). 
  
  \textbf{Quality of the synthetic images}. Although the generator can restore low-resolution images, limited by the generator's performance, the synthetic samples with high resolution are still a little ambiguous. For some tasks that require high resolution, such as fine-grained vision classification challenges, it may bring adverse effects.
  
  \textbf{Non-convolution Networks}. In our approach, BNS is a regularization item for optimizing synthetic images, since the distribution information~(i.e., mean and variance) of BN has been pre-defined and stored in the parameters of the model during the pre-training process.
  Therefore, it can be exploited in inversion and optimization tasks. However, in some non-convolution networks such as transformer networks, the layer normalization is used, which does not contain distribution information.
  As such, they may be less effective in DFKD.

\subsection{Feature work}

We suggest some future work to solve the limitations explained above:
    
    \textbf{Improve distillation efficiency}. In addition to the training of DFKD by pre-training and fine-tuning, we suggest the following two aspects to improve the training efficiency of the model. Firstly, for the student model, the different knowledge forms of the teacher model from the intermediate layers, such as relationship and attention, etc., can be comprehensively utilized to improve the distillation efficiency of the model. Second, for the generator, the model can be improved by optimizing the generation process, such as extracting knowledge from previously synthesized images. 
     
    \textbf{Improve synthetic image quality}. For complex tasks, the quality of synthetic images is crucial for downstream distillation. We recommend combining super-resolution learning or image denoising-related domain knowledge to improve image quality.
    This could make the proposed method robust in more challenging scenarios. 
    
    \textbf{Other optimization information}. For the non-convolution networks, we consider combining the decoder architecture with other optimizations such as gradient information to benefit synthetic images.
    This might be beneficial for boundary fitting for the studied task.

\section{Acknowledgements}
This research was supported by National Natural Science Foundation of China under Grants 92270119, 62072182, and 62272165.















\bibliographystyle{elsarticle-num}
\bibliography{ref}






\end{document}